\documentclass[runningheads]{llncs}
\usepackage{graphicx}
\usepackage{comment}
\usepackage{amsmath,amssymb} 
\usepackage{color}
\usepackage[table]{xcolor}
\usepackage{booktabs}
\usepackage{textcomp}

\begin{document}
\pagestyle{headings}
\mainmatter
\def\ECCVSubNumber{2744}  

\title{FAIRS – Soft Focus Generator and Attention for Robust Object Segmentation from Extreme Points} 


\titlerunning{FAIRS for Robust Object Segmentation from Extreme Points}
%
\author{Ahmed H. Shahin\inst{1}\thanks{Work done during an internship at Inception Institute of Artificial Intelligence.} \and
Prateek Munjal\inst{2}\and
Ling Shao\inst{3} \and Shadab Khan\inst{2}}
\authorrunning{A. Shahin et al.}
%
\institute{Centre for Medical Image Computing, University College London, London, UK\\
\email{a.shahin@cs.ucl.ac.uk}\\
\and
Group42 Healthcare, Abu Dhabi, UAE\\
\email{\{prateekmunjal31,skhanshadab\}@gmail.com}
\and
Inception Institute of Artificial Intelligence, Abu Dhabi, UAE\\
\email{ling.shao@inceptioniai.org}}
\maketitle

\begin{abstract}
Semantic segmentation from user inputs has been actively studied to facilitate interactive segmentation for data annotation and other applications. Recent studies have shown that extreme points can be effectively used to encode user inputs. A heat map generated from the extreme points can be appended to the RGB image and input to the model for training. In this study, we present FAIRS – a new approach to generate object segmentation from user inputs in the form of extreme points and corrective clicks. We propose a novel approach for effectively encoding the user input from extreme points and corrective clicks, in a novel and scalable manner that allows the network to work with a variable number of clicks, including corrective clicks for output refinement. We also integrate a dual attention module with our approach to increase the efficacy of the model in preferentially attending to the objects. We demonstrate that these additions help achieve significant improvements over state-of-the-art in dense object segmentation from user inputs, on multiple large-scale datasets. Through experiments, we demonstrate our method's ability to generate high-quality training data as well as its scalability in incorporating extreme points, guiding clicks, and corrective clicks in a principled manner.
\end{abstract}

\section{Introduction}
Semantic segmentation has been one of the longstanding problems in computer vision. Segmentation algorithms produce masks to classify the pixels into foreground/background classes. These algorithms are used for a wide variety of tasks, ranging from typical applications in security \cite{Snidaro2005VideoIntelligence}, robotics \cite{Milioto2019Bonnet:CNNs}, satellite imaging \cite{Bhandari2014CuckooEntropy}, medical imaging \cite{Ronneberger2015U-Net:Segmentation}, to other interesting applications such as counting number of penguins in their arctic colonies \cite{Arteta2016CountingWild}. Such algorithms require a large amount of ground truth labeled data for training, which is annotated with human oversight and is therefore slow and expensive. To reduce the costs and accelerate the annotation process, methods to generate annotations from user inputs have been widely studied \cite{Papadopoulos2017ExtremeAnnotation,Maninis2018DeepSegmentation}.

\begin{figure}[t]
\centering
\includegraphics[width=0.99\linewidth]{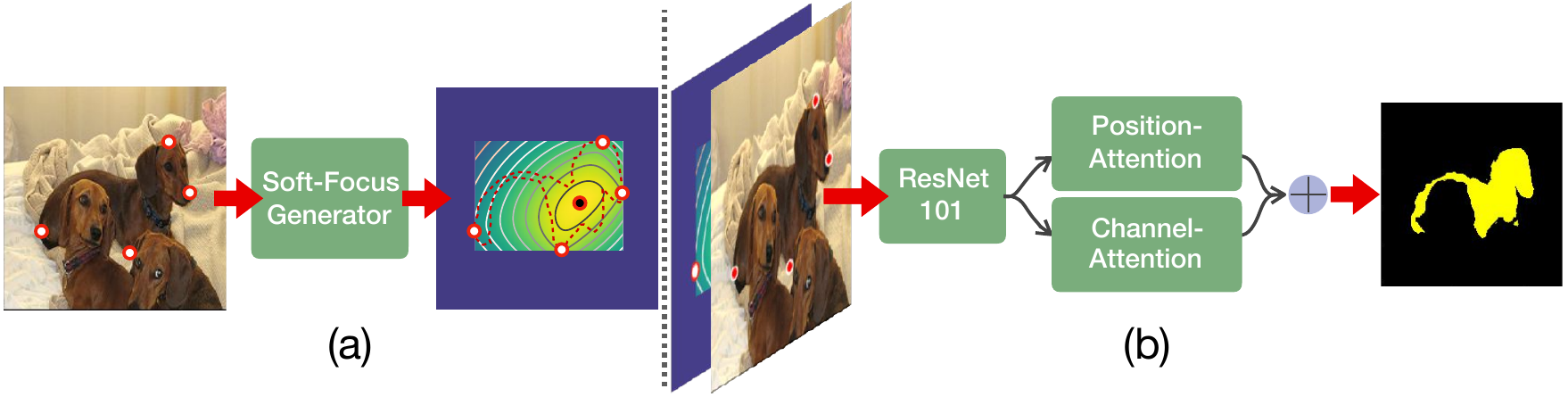}
\vspace{-2mm}
\caption[]{Overview of our approach to generate object segmentation from extreme points using the proposed Soft-Focus Generator (SFG) module that results in a nearly-convex smoothly varying potential field using an $n$-ellipse formulation as shown in (a). (b) shows overview of the rest of the pipeline, we use ResNet-101 backbone with a dual attention module as proposed in \cite{fu2019dual} to produce a segmentation of the object of interest.}
\label{figure1}
\vspace{-3mm}
\end{figure}

Several promising methods have been proposed that rely on user-provided cues such as bounding box \cite{Dai2015BoxSup:Segmentation}, clicks \cite{Maninis2018DeepSegmentation,Li2018InteractiveDiversity}, and scribbles \cite{Lin2016ScribbleSup:Segmentation}. These methods have worked to a varying degree of success on various datasets, and incorporating such cues from the users in a principled manner that works across datasets and conditions remains an open problem. One particular form of user clicks, called extreme points (EPs), have recently received significant attention owing to the study by \cite{Papadopoulos2017ExtremeAnnotation}, which showed that extreme points can be annotated more quickly than other forms of user inputs such as a bounding box.

This study, proposes a principled approach to encoding information from extreme points and corrective clicks using a Soft Focus Generator (SFG) that produces a heat map, which is input to the model for generating a dense segmentation mask (Figure \ref{figure1}). Further, equipped with a dual-attention module, our approach generates high-quality segmentation masks on a variety of challenging datasets such PASCAL \cite{EveringhamM.andVanGoolL.andWilliamsC.K.I.andWinnJ.andZissermanTheResults}, COCO \cite{Lin2014MicrosoftContext}, SBD \cite{hariharan2011semantic}, GrabCut \cite{Rother2004GrabCutCuts}, and Berkeley \cite{McGuinness2011TowardSegmentation}. Compared to several state-of-the-art approaches on object segmentation from user inputs, our method (FAIRS – Focus and Attention for Interactive Robust Segmentation) took fewer clicks to achieve superior evaluation metrics on all comparative experiments conducted in this study.

We demonstrate FAIRS's effectiveness in generating training data through an exacting experiment where we trained a previously-untrained version of our model using only the synthetic labels generated by a trained version of our model. This weakly-supervised version of FAIRS achieves results that are at-par with the state-of-the-art approaches in object segmentation from user inputs, and only lags behind the version of FAIRS trained using ground truth labels. Further, we also evaluate FAIRS's performance when presented with lower than 4 extreme points or when presented with corrective clicks for refinement during interactive segmentation. We find that FAIRS handles these diverse scenarios very well, while maintaining annotation quality.

\section{Related Work}
\noindent\textbf{Segmentation from User Inputs}
\label{extreme_points}
In recent literature, segmentation from user-provided cues such as bounding box \cite{Xu2017DeepSelection}, patches \cite{pinheiro2015learning,pinheiro2016learning}, scribbles \cite{Lin2016ScribbleSup:Segmentation}, and clicks \cite{Papadopoulos2017ExtremeAnnotation,Acuna2018EfficientPolygon-RNN++,hu2019fully}, have been investigated. In state-of-the-art in segmentation from user inputs, Liew \textit{et al.} \cite{liew2019multiseg} use an image and a distance transform map computed from user input to produce multiple segmentation masks that are fed to the computationally expensive post processing step of non-maximum suppression and graph cut. Jang \textit{et al.} \cite{jang2019interactive} encode the user inputs using distance map and input it along with the image to an FCN architecture; their method relies on 10 iterations of forward-backward propagation for refining the output.

Further, Majumder \textit{et al.} \cite{majumder2019content} augment the RGB channels by adding 4 additional channels -- 2 for superpixel based guidance, and 1 channel each for object-based guidance and distance transform. Lin \textit{et al.} \cite{lin2019block} propose a block annotation module for online annotation that asks users to annotate blocks iteratively until a satisfactory segmentation is achieved. Li \textit{et al.} \cite{Li2018InteractiveDiversity} compute positive and negative distance maps which are appended to activation maps extracted from an FCN pipeline fed to a segmentation model that produces multiple segmentation masks that are post-processed through a selection network for final output.

Lastly, extreme points (EPs) have also been used for segmentation from user input \cite{Maninis2018DeepSegmentation}. Maninis \textit{et al.}. \cite{Maninis2018DeepSegmentation} presented deep extreme cut (DEXTR), where a 2D heat map is computed using EPs, with the purpose of guiding the network to the object-of-interest. Wang \textit{et al.} built on the ideas presented in \cite{Maninis2018DeepSegmentation}, by developing a post-processing module that combines image features extracted from CNN with a level set extraction method (DELSE) \cite{Wang2019ObjectEvolution} for refinement. EPs were also used in a full-image segmentation study by Agustsson \textit{et al.} \cite{Agustsson2019InteractiveJointly}, who proposed a method that takes four EPs per object and variable number of corrective scribbles in an image to produce an image-level segmentation mask, though in contrast to \cite{Wang2019ObjectEvolution,Maninis2018DeepSegmentation}, they used a 6 pixel wide circle to represent the extreme points.

The studies cited above have attempted to incorporate rich spatial information as a guidance by relying on sophisticated pre-processing \cite{majumder2019content} and/or computationally expensive post-processing steps\cite{Wang2019ObjectEvolution}, which is not ideal for fast interactive segmentation. Further, while the methods using extreme points \cite{Wang2019ObjectEvolution,Maninis2018DeepSegmentation,Agustsson2019InteractiveJointly} work well, the approach of placing only small-width (=10 px) Gaussians or circles at EPs does not make any other distinction between foreground/background classes, since most of the background and foreground pixels (\texttildelow99.85\% at 512$\times$512 resolution) are placed at nearly zero weight. This leads to issues with segmentation output for objects that may be occluded or have confusing textural properties, such as presence of texture-less high contrast patches (e.g. dalmatian dog), or suffer from annotation error. This avenue of improvement has inspired recent methods, though finding the best method to encode the background/foreground spatial information from user inputs remains an open problem.

\noindent\textbf{Contribution}: In this study, we set out to design a mechanism for improving guidance to the neural networks. We accomplish this in two ways. 
First, starting from the input end of the network, we propose a novel mechanism to incorporate a simple and scalable distance map by using the $n$-ellipse (also called multi-foci ellipse) formulation. Second, at the output end of the network, we integrate a dual-attention module as proposed in \cite{fu2019dual} within our pipeline to encourage network's preferential attention to the foreground. While attention modules have been used to improve performance of a segmentation model where object classes are known, their efficacy on class-agnostic segmentation has so far not been reported. We further note that while the attention maps are part of the architecture and are learnt, $n$-ellipse heat map provide a computed (not learnt) diffused focus as an additional input to the network. Through a series of experiments we comprehensively demonstrate the effectiveness of our approach in producing high-quality segmentation masks as well as its ability to work with a combination of extreme points and corrective clicks. In the process, we achieve performance superior to several state-of-the-art methods on object segmentation from user inputs on multiple challenging datasets.

\section{Methods}
\subsection{Soft Focus Generator (SFG)}
\label{nellipse}
The SFG comprises primarily three computation steps. First, we compute the potential field of an $n$-ellipse map. Subsequently, we post-process the potential field to generate the soft focus map and crop it with the bounding box defined by extreme points. When corrective clicks are added, they are compounded with the cropped soft focus map at this stage. Finally, we compute the Gaussians heat map as used by \cite{Wang2019ObjectEvolution,Maninis2018DeepSegmentation,Agustsson2019InteractiveJointly} and merge it with the soft focus map. These individual steps are described next.
\vspace{2mm}

\begin{figure}[!b]
\centering
\includegraphics[width=0.99\linewidth]{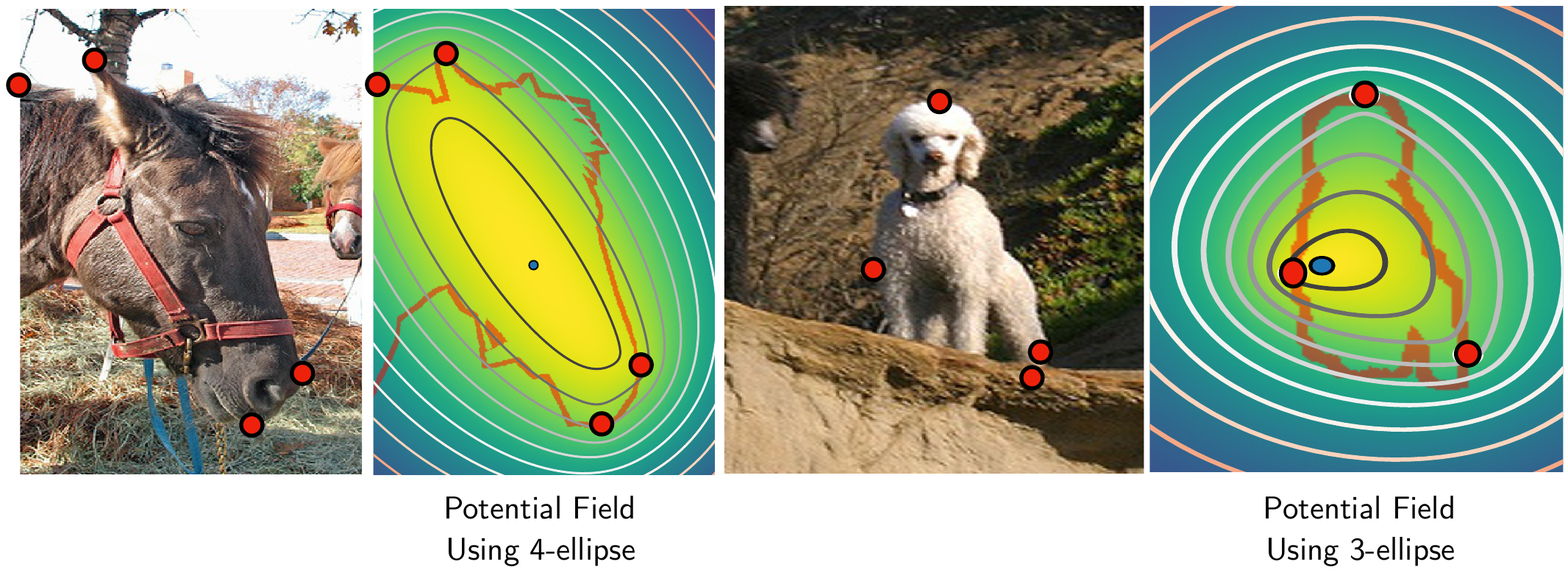}
\caption[]{An illustration of post-processed potential field calculated over the image using three and four extreme points. The contrast has been exaggerated for visualization. The potential field places a soft focus on the object that slowly decays away from towards the background.}
\label{figure_nellipse}
\end{figure}

\noindent\textbf{1. $n$-ellipse potential field – $\pi$}: To formulate $\pi(\cdot)$, we began with the goal of achieving the following properties in the soft focus map: (i) it should be simple and fast to compute, (ii) it should scale well with the number of points, (ii) it should encode spatial relationship between the extreme points. Multi-focal ellipses or $n$-ellipse, which generalize a simple ellipse (with two focal points) to higher number of focal points, fit all of these desired properties figure \ref{figure_nellipse}.

Provided with an image $I(x) \in \Omega; x \in \mathbb{R}^2$ and $n$ focal points of an object of interest $p _n= {p_1, p_2, ..., p_i}; $ $p_i \in \mathbb{R}^2$, we compute the potential field $\pi(p_n,x)$ using: $\pi(p_n, x)=\sum_{i=1}^n \left \| x-u_i \right \|_{2}$. By definition, a 1-ellipse is a typical circle, and a 2-ellipse is an ellipse. This potential field is a smoothly varying distribution of weights over the image with nearly-convex isocontours. Using the $n$-ellipse formulation enables us to use variable number of extreme points, not necessary four, while preserving a consistent smooth assignment of weights to the foreground region.
\vspace{2mm}

\noindent\textbf{2. Post-processing $\pi(p_n,x)$}: Once the potential field is calculated, we transform it using:
\begin{equation}
    \hat{\pi}(p_n,x) = N\left ( \frac{1}{\pi(p_n,x)} \right )^{\beta} \circ B(x)
\end{equation}
In this equation, $N(\cdot)$ normalizes the range of its argument to 0–1, $\beta$ is a hyper-parameter that controls the potential decay rate from the center of the $n$-ellipse, and $B(x): \mathbb{R}^2\to \{0,1\}$ is a mask that is 0 outside of the bounding box and 1 everywhere inside. Limiting $\hat{\pi}(p_n,x) $ to the extent of bounding box adds an implicit cue about background pixels. Figure \ref{figure_nellipse} visualized potential field for a couple of images with 3 and 4 extreme points.
\vspace{2mm}

\noindent\textbf{3. Incorporating corrective clicks}: For interactive segmentation, corrective clicks can be incorporated over $\hat{\pi}(p_n,x) $ by composing Gaussians for false-positives-corrective (FPC) click as well as false-negative-corrective (FNC) click. The FPC clicks at $x_{fpc}^i; i=1...n$ are encoded using a Gaussian heat map $g(x_{fpc})$, likewise, FNC clicks are encoded using 1-$g(x_{fnc})$. The FPC and FNC maps are compounded with $\hat{\pi}(p_n,x) $ to produce  $\tilde{\pi}(p_n,x)$ using the equations below:
\begin{equation}
\breve{\pi}(p_n,x) = 
\begin{cases}
    \hat{\pi}(p_n,x,x_{fnc}),& \text{if }  \hat{\pi}(p_n,x)< 1-g(x_{fnc}) \\
    1-g(x_{fnc}),              & \text{otherwise}
\end{cases}
\end{equation}
\vspace{-3mm}\begin{equation}
\tilde{\pi}(p_n,x, x_{fpc}, x_{fnc}) = \text{max}(\breve{\pi}(p_n,x), g(x_{fpc}))
\end{equation}
We note that when the corrective clicks are not provided, as is the case for segmentation using four extreme points, $\tilde{\pi}(p_n,x, x_{fpc}, x_{fnc}) =\hat{\pi}(p_n,x) $.
\vspace{2mm}

\noindent\textbf{4. Incorporating extreme points}:

The soft focus map $\psi(p_n,x, x_{fpc}, x_{fnc})$ is computed by compounding Gaussians placed at extreme points with the post-processed potential field using: \\ $\psi(p_n,x, x_{fpc}, x_{fnc}) = \text{max}(\tilde{\pi}(p_n,x, x_{fpc}, x_{fnc}), g(p_n))$.

\begin{figure}[!b]
\centering
\includegraphics[width=0.99\linewidth]{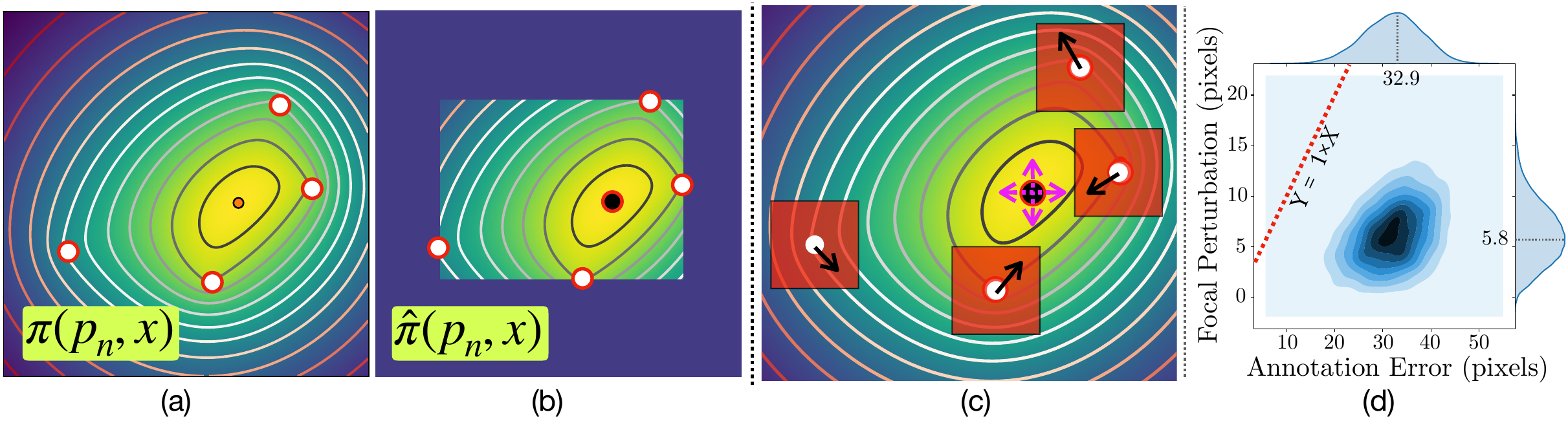}
\caption[]{Intermediate outputs of SFG are shown in (a) and (b). (c) illustrates an experiment done to assess the robustness of SFG to annotation error, the red boxes indicate the extent of simulated annotation error (-10--10 px), and the resulting induced perturbation on the focal point is measured. (d) show density plot of annotation error vs focal perturbation, we observe that focal point moves relatively little compared to the annotated extreme points.}
\label{figure_SFG}
\end{figure}

\noindent The intermediate outputs of SFG are shown in \ref{figure_SFG}.

\subsection{Network Design}
\label{attention}
Similar to \cite{Maninis2018DeepSegmentation,chen2017deeplab,chen2017rethinking,chen2018encoder,fu2019dual} we begin with a ResNet-101 backbone \cite{He2016DeepRecognition} equipped with dilated convolutions, and add a positional and spatial attention modules \cite{fu2019dual}, to model the semantic relations in spatial and channel dimensions. We remove the down-sampling operations and employ dilated convolutions in the last two blocks to preserve a reasonable spatial resolution (1/8 of the original image resolution).

Two parallel attention heads are applied to the output feature maps. Spatial attention head generates a map in which each pixel is a weighted sum of all pixels in the feature maps. Thus, it encodes global information and promotes the parts that are semantically related to the object blobs. Similarly, channel attention head models channel correlations by learning to promote channels that are relevant to the object segmentation task. The output of the two attention blocks is fused and used to produce the final segmentation mask.

\section{Experiments and Results}
We extensively experimented using FAIRS on five public datasets: Berkeley segmentation dataset \cite{McGuinness2011TowardSegmentation}, PASCAL 2012 \cite{EveringhamM.andVanGoolL.andWilliamsC.K.I.andWinnJ.andZissermanTheResults}, GrabCut \cite{Rother2004GrabCutCuts}, COCO \cite{Lin2014MicrosoftContext}, and SBD. In this section, we first discuss the model implementation and training details, followed by details of datasets and the experiments and results. We discuss multiple experiments on using FAIRS for class-agnostic segmentation, including realistic evaluation on human-annotated extreme points, generalization across dataset, and generalization to seen and unseen object categories. Further, we discuss results from experiments where we use annotations generated by our model to train a weakly-supervised version of FAIRS, and demonstrate FAIRS's ability in generating high-quality annotations. Lastly, we discuss our results on using FAIRS for interactive segmentation and demonstrate the flexibility of our approach in achieving outstanding results with different number of clicks and corrective clicks in interactive mode.

\subsection{Model Training and Datasets}
\noindent\textbf{Model Training:}\hspace{1mm} As is standard practice with segmentation architectures using ResNet backbone \cite{He2016DeepRecognition}, we initialize our model's backbone using pre-trained ImageNet weights \cite{Russakovsky2015ImageNetChallenge}. In order to work with pre-trained weights, we copy the $3^{rd}$ channel kernel-weights to the $4^{th}$ channel in the input layer. Attention module was initialized randomly. We used a multi-term loss with equal weights to train the model, where each term is a weighted cross entropy to alleviate the class imbalance. We use random scaling, rotations and horizontal flips for augmenting our dataset. We use SGD with momentum as the optimizer to train our model. We computed our heat map by setting $\beta=5$, $\kappa=2$ and Gaussian $\sigma=10$. We compute our object masks on all images, including multiple objects within each image. Dataset-specific training details are mentioned next.

\vspace{2mm}
\noindent\textbf{COCO \cite{Lin2014MicrosoftContext}:}\hspace{1mm} We train using 82783 number of images from 2014 Coco train containing 80 object classes. We test on COCO 2017 validations set (\texttildelow5k images, \texttildelow36k objects). We train the model using a learning rate $1\times10^{-7}$, batch size of $48$, and for 15 epochs due to the large number of images in the dataset.

\vspace{1mm}
\noindent\textbf{PASCAL \cite{EveringhamM.andVanGoolL.andWilliamsC.K.I.andWinnJ.andZissermanTheResults,hariharan2011semantic}:} \hspace{1mm} We train using PASCAL data in two different ways, one with PASCAL and SBD data (10582 images), and one with only PASCAL data (1464 images). We make the distinction between two versions of our model where necessary, otherwise, PASCAL, should be taken to mean PASCAL and SBD (10582 images) as this is the common practice when referring to this dataset. We train on PASCAL with an initial constant learning rate of $1\times10^{-7}$ for 100 epochs, and then reduce it to $5\times10^{-8}$ and train for another 50 epochs.
\vspace{1mm}\\
\noindent\textbf{Berkeley \cite{martin2001database}:} We do not train using Berkeley train set. We test on 100 object masks extracted from 96 images, provided by \cite{McGuinness2010AAlgorithms,jang2019interactive}.
\vspace{1mm}\\
\noindent\textbf{GrabCut: \cite{Rother2004GrabCutCuts}} We evalute our model on GrabCut's test set (50 images).
\vspace{1mm}\\
\noindent\textbf{SBD \cite{hariharan2011semantic}:} For reporting on SBD, we trained only on SBD data (8498 train images) for fair comparison with other methods. We test on SBD validation set (2820 images, all objects). We trained with an initial constant learning rate of $1\times10^{-7}$ for 100 epochs, and then reduce it to $5\times10^{-8}$ and train for 25 epochs.
\vspace{1mm}\\
\noindent\textbf{User Input:}\hspace{1mm} We follow the approach used by \cite{Maninis2018DeepSegmentation}, and infer extreme points by extracting them from the ground truth mask. To simulate noise in the extreme points, we add uniformly distributed noise of 10 px to their coordinates.

\subsection{Ablation Study}
We evaluated the relative gains by incorporating dual attention module and the output of soft focus generator (SFG). We adopt the DeepLab-V2 \cite{chen2017deeplab} with ResNet-101 backbone and PSP head \cite{Zhao2017PyramidNetwork,Maninis2018DeepSegmentation} as the base. We do not use the PSP head with the attention module. We performed the ablation study using PASCAL-train and PASCAL+SBD-train data, and observed that, both attention module and our heat map improved the IoU scores as shown in table \ref{ablation}.
\begin{table}[b]
\centering
\begin{tabular}{lll}
\toprule
Model  & PASCAL                                  & PASCAL+SBD\\
\midrule
Base Model & 90.50\%                & 91.50\% \\
+ Dual Attention & 91.22\% &  91.80\%  \\
+ SFG & \textbf{91.56\%} & \textbf{92.22\%} \\
\bottomrule
\vspace{-1mm}
\end{tabular}
\caption{Results from ablation study. We observed a gain in IoU with dual attention module as well as $n$-ellipse heat map.}
\label{ablation}
\end{table}

\begin{figure}[t]
\centering
\includegraphics[ width=0.99\linewidth]{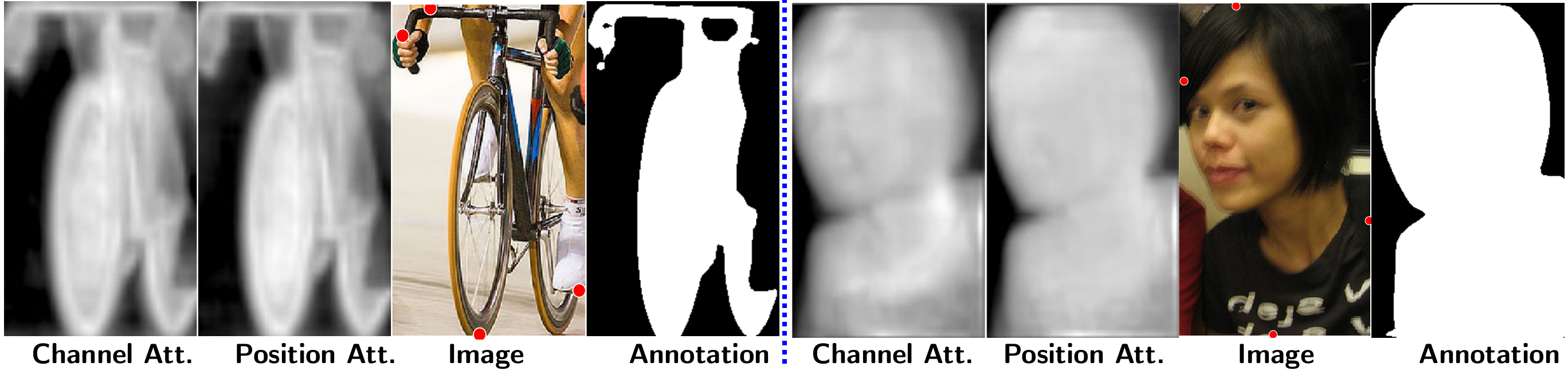}
\caption[]{Attention maps generated for two different images are shown. On closer inspection we observe that channel attention module is attending to finer details, e.g. contours of the face, nose, rim of the cycle wheel etc, whereas the position attention module is attending preferentially to the foreground.}
\label{figure_attention}
\vspace{-3mm}
\end{figure}

\vspace{1mm}
\noindent \textbf{Comparison with DEXTR}: To further assess the utility of SFG module, we replaced the Gaussian heat map in DEXTR pipeline with the output of SFG, and trained the resulting model on PASCAL+SBD-train dataset. Compared to DEXTR's 91.50$\%$ IoU, DEXTR with SFG achieves 91.81\% IoU.
\vspace{1mm}

\noindent \textbf{Why does SFG work?} DEXTR places small Gaussians at extreme points (EP), this is problematic if the contrast around EP is low or if annotations are imperfect, as the texture cues around EP can be misleading. To this end, many studies \cite{jang2019interactive,Li2018InteractiveDiversity,liew2019multiseg,majumder2019content}, have used distance transforms as one of the cues. Both Gaussians and distance transforms have peaks (+ve/-ve) at the annotated points (e.g. \cite{liew2019multiseg} fig.2, \cite{jang2019interactive} fig.1, \cite{majumder2019content} fig.1) and distribute density around these peaks, thus suffer from annotation error or low texture around EPs. Further, errors in annotation can aggravate the problem as all of these methods place multiple peaks in the heat map. We conjecture SFG overcomes these issues because it uses $n$-ellipse potential field which has unimodal peak of the density on object foreground, rather than EPs and is therefore robust to annotation error. To test the robustness of SFG, we did an experiment where we perturbed both the \textit{x} and \textit{y} index of all 4 EPs using a random number drawn from a uniform distribution to simulate annotation error ($r{\raise.17ex\hbox{$\scriptstyle\mathtt{\sim}$}}
U[-10,10]$), and measured the perturbation induced on the focus of $n$-ellipse potential field. As indicated by the marginals in the density plot (fig. \ref{figure_SFG}d, 10000 draws) of annotation error vs induced perturbation, for a mean annotation error $\approx 32.9$ px, the mean induced perturbation on the focal point is $\approx 5.8$ px. These results support our hypothesis. Qualitative results using our pipeline with SFG vs our pipeline with Gaussians (DEXTR's approach) have been shown in the supplementary material, where we also present example cases where FAIRS does not improve over DEXTR.

\begin{figure}[!ht]
\centering
\includegraphics[width=0.74\linewidth]{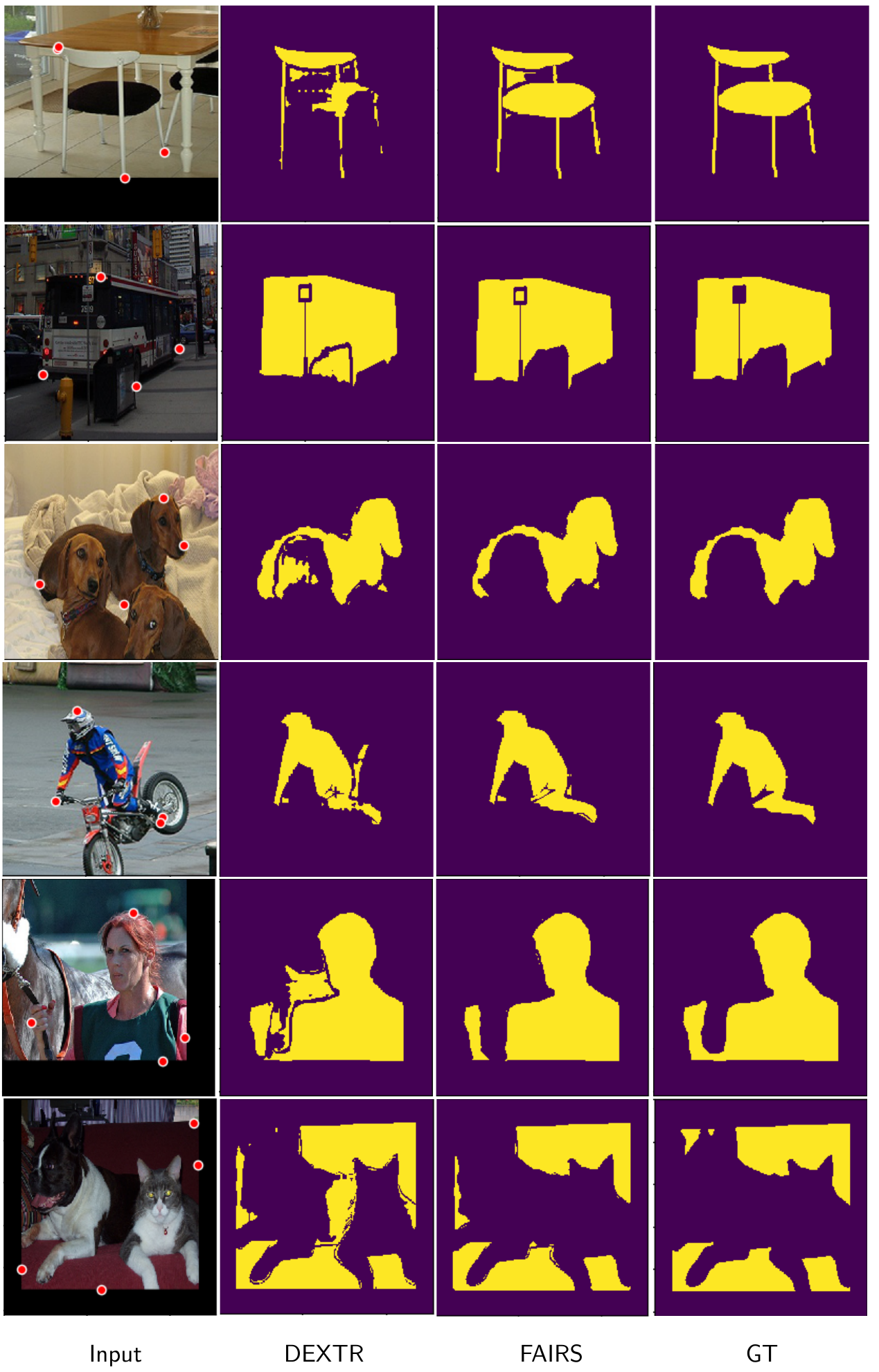}
\caption[]{We demonstrate FAIRS' ability to deal with a variety of challenging scenarios in comparison to DEXTR. The network is able to preferentially attend to foreground with the help of attention modules and soft focus map, which helps it perform under challenging conditions. For example, all extreme points of the chair are on the white patch, yet, FAIRS was able to produce a reliable segmentation with a small false +ve blob that can be further refined using corrective clicks. Similarly, in the second row, FAIRS is able to recognize that the newspaper dispenser is not a part of the bus, despite one of the extreme points being close to it. In the last row image, FAIRS is able to deal with presence of texture-less contrast in a better-controlled manner compared to DEXTR. Additional results in supplementary material.}
\label{figure_panel}
\end{figure}

\noindent \textbf{Attention Module}: To qualitatively assess the output of the attention modules, we visualize their activations in figure \ref{figure_attention}. We find that although both the attention modules, have similar overall structure, on taking a closer look, we find that channel attention module seems to be focusing on finer details, whereas position attention module is attending to the coarser foreground. This nature of attending to different details helps improve performance on challenging cases, figure \ref{figure_panel}.

\subsection{Class-Agnostic Segmentation}
FAIRS can be used for class-agnostic segmentation, using extreme points as cues provided by the user. This object can be of any class, and it can be different than classes present in the training set. We perform a number of experiments to benchmark our method's performance on class-agnostic segmentation tasks, including generalization to unseen datasets and unseen classes.

\vspace{3mm}
\noindent\textbf{Human-Annotated Extreme Points:}\hspace{1mm} We use the extreme points provided by \cite{Papadopoulos2017ExtremeAnnotation}, covering a subset of PASCAL+SBD train-set images, for evaluating FAIRS under realistic conditions. Human-annotated extreme points collection was crowd-sourced on 5623 images by \cite{Papadopoulos2017ExtremeAnnotation}. To be consistent with \cite{Maninis2018DeepSegmentation}, we refer to this dataset as $\text{PASCAL}_{\text{EXT}}$. We predict segmentation masks using FAIRS trained on COCO and calculate the IoU. The results are shown in the Table \ref{real_points}, FAIRS outperforms all other methods on this evaluation.

\begin{table}[h]
\parbox{.45\linewidth}{
\centering
\begin{tabular}{lc}
\toprule
Method                                    & IoU    \\
\midrule
Sharpmask from bounding box \cite{Papadopoulos2017ExtremeAnnotation} & 69.3\% \\
GrabCut using extreme points \cite{Papadopoulos2017ExtremeAnnotation} & 73.6\% \\
Sharpmask upper bound & 78.0\%\\
DEXTR from extreme points\cite{Maninis2018DeepSegmentation}         & 80.1\% \\
FAIRS (\textbf{Ours}) from extreme points          & \textbf{84.0\%}     \\
\bottomrule
\vspace{-1mm}
\end{tabular}
\caption{FAIRS (trained on COCO objects dataset) compared to other methods on class-agnostic segmentation from human-annotated extreme points on $\text{PASCAL}_{\text{EXT}}$.}
\label{real_points}
}
\hfill
\parbox{.45\linewidth}{
\centering
\begin{tabular}{lccc}
\toprule
Dataset & DELSE & DEXTR & Ours \\
\midrule
COCO         & -- & 87.8\% & \textbf{90.6\%} \\
PASCAL       & 90.5\% & 90.5\% & \textbf{91.5\%} \\
\begin{tabular}{@{}c@{}}PASCAL \\ + SBD\end{tabular} & 91.3\% & 91.5\% & \textbf{92.2\%} \\
\bottomrule
\vspace{-1mm}
\end{tabular}
\caption{Three different models were trained on large multi-class segmentation datasets using simulated extreme points. Resulting IoU scores on PASCAL 2012 validation set are shown. \vspace{0.45cm}}
\label{seg_comparison}
}
\end{table}
Table \ref{real_points} shows that the IoU using FAIRS is significantly better than DEXTR, GrabCut-based approach and sharpmask. This demonstrates our method's ability to generalize well to human-provided extreme points despite being trained on simulated extreme points.

\begin{figure}[!t]
\centering
\includegraphics[width=0.75\linewidth]{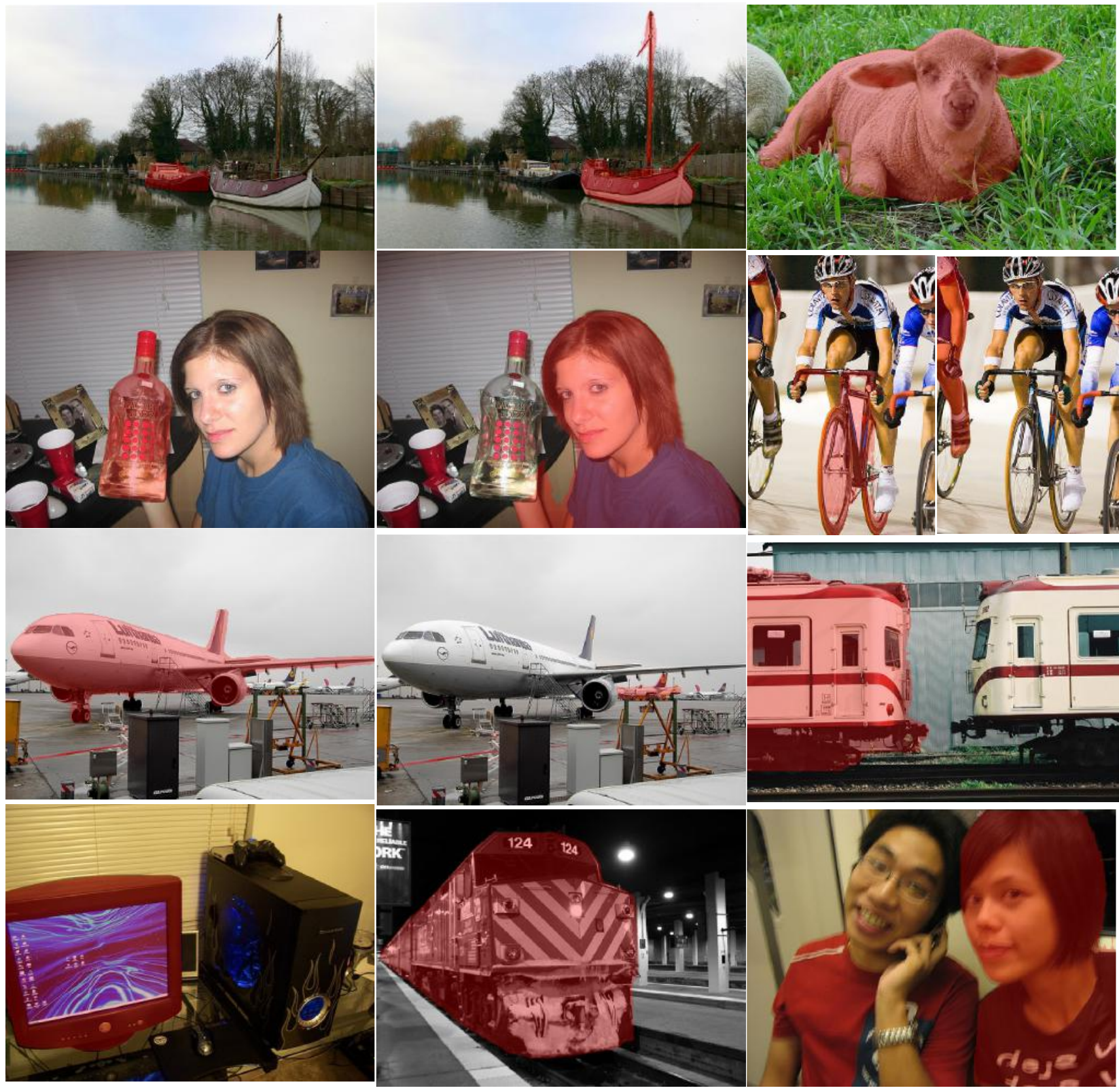}
\caption[]{A few example masks predicted using FAIRS shown overlaid on images from PASCAL dataset. We particularly highlight FAIRS's robustness to objects surrounded by clutter, e.g. top row middle image of the boat, and 2nd row right most image of the bicycle. Additional results in supplementary material.}
\label{figure_FAIRS}
\end{figure}

\vspace{3mm}
\noindent\textbf{Segmentation From Simulated Extreme Points:}
We experiment with large scale datasets by simulating extreme points as described previously. We trained three versions of FAIRS for this study. The three models were trained using the training dataset from COCO 2014 train set, PASCAL 2012 train set (1464 images), and PASCAL 2012 train set combined with SBD train set (10582 images). In table \ref{seg_comparison} below, we compare the IoU score on PASCAL 2012 validation set. Qualitative comparison of masks achieved with FAIRS and DEXTR are shown for a number of challenging cases in figure \ref{figure_panel} and demonstrate FAIRS's robustness. Results in table \ref{seg_comparison} show that our method improves significantly over other state-of-the-art methods. Further in figure \ref{figure_FAIRS}, we show FAIRS's ability to deal with cluttered and non-cluttered scenes for object segmentation using SFG output computed from four extreme points.

\vspace{4mm}
\noindent\textbf{Comparison to State-of-The-Art and Number of Clicks:}\hspace{1mm} In order to compare with state-of-the-art methods that offer segmentation from user-inputs, we trained FAIRS, 3 and 4 extreme points, in addition to upto 2 corrective clicks. Our method of simulating 3 points, and corrective clicks is described as follows.

We simulate three extreme points by: (i) obtaining 4 extreme points, (ii) identifying a pair of extreme points that are closest to each other, (iii) dropping one of the points in the pair randomly during training. This closely simulates the actual use-case where an annotator would preferentially select extreme points such that the coverage of the objects is maximized. Further, corrective clicks were simulated by identifying largest false positive and false negative blobs, and sampling a point randomly from the largest blob across the false positives and negatives. We note that soft focus generator was able to handle all of these scenarios automatically for producing the soft focus map. 

To demonstrate the efficiency of our encoding method with lower number of extreme points, we report the number of clicks needed by each algorithm to reach certain IoU on Berkeley, PASCAL validation set, GrabCut, and SBD validation set. The results organized by dataset are shown in the table \ref{sota_comparison}. FAIRS outperforms all state-of-the-art methods that we compared with on all of the datasets that we experimented with.
 \cite{Maninis2018DeepSegmentation}

\begin{table*}[!t]
\centering
\begin{tabular}{lcccc}
\toprule
Methods\textbackslash Datasets & PASCAL 85\% & Berkeley 90\% & GrabCut 90\% & SBD 85\% \\
\midrule
RIS-Net \cite{Liew2017RegionalNetworks}                        & 5.7          &  --              & 6.0          &     --      \\
Latent Diversity  \cite{Li2018InteractiveDiversity}                    &   --           &  --              & 4.79         & 7.41      \\
DEXTR \cite{Wang2019ObjectEvolution}                         & 4.0(91.5\%)  & 4+ (89.1 @4)   & 4.0          &   --        \\
CAMLG  \cite{majumder2019content}                         & 3.62         & 5.6            & 3.5          &  --          \\
FCTSFN  \cite{hu2019fully}                        & 4.58         & 6.49           & 3.58         &  --         \\
MultiSeg    \cite{liew2019multiseg}                    & 3.51         & 4.00           &  --            &  --          \\
BRS-DenseNet \cite{jang2019interactive}                   &   --           & 5.08           & 3.60         & 6.59      \\
\textbf{FAIRS-WS} (\textbf{Ours})       & 4.0 (91.4\%)     & 4.0 (88.8\%) & 4.0 (92.8\%) &    --         \\
\textbf{FAIRS (Ours)}       & \textbf{3.0} (\textbf{88.9\%}) & \textbf{4.0} (\textbf{91.9\%})   & \textbf{3.0} (\textbf{91.9}\%)   &  \textbf{4.0} (\textbf{88\%}) \\
\bottomrule
\vspace{-1mm}
\end{tabular}
\caption{We compare FAIRS's effectiveness with state-of-the-art methods on segmentation from user inputs on multiple datasets. Resulting IoU scores are shown. Most recent methods are in the last few rows of the table. We note that FAIRS-WS was trained using only the generated masks on COCO 2017 validation set using FAIRS trained on PASCAL. FAIRS-WS result demonstrate the effectiveness of generated labels in actual training for a realistic use-case.}
\label{sota_comparison}
\end{table*}

\vspace{3mm}
\noindent\textbf{Generalization to Unseen Classes:}\hspace{1mm} We evaluate FAIRS's ability to generalize to unseen classes by training the model on PASCAL training set, and evaluating on COCO 2017 validation set. In this experiment, we run FAIRS to compute masks for object classes that are not present in the PASCAL train set (COCO Unseen), this results in 60 object classes, with \texttildelow15k number of objects. We compute masks for all objects. The results are shown in table \ref{generalization_unseen}, which shows that FAIRS suffers from a negligible comparative performance drop when we use it to segment object classes that were absent from the training data.
\begin{table}[b!]
\centering
\begin{tabular}{llll}
\toprule
Train  & Test                                  & DEXTR  & Ours \\
\midrule
PASCAL & COCO Seen        & 80.3\% &  \textbf{81.8\%}  \\
PASCAL & COCO Unseen & 79.9\% &   \textbf{81.7\%} \\
\bottomrule
\vspace{-1mm}
\end{tabular}
\caption{Evaluation of FAIRS's ability to generalize on unseen classes of COCO 2017 Validation set.}
\label{generalization_unseen}
\end{table}

\vspace{3mm}
\noindent\textbf{Generalization to Unseen Datasets: } In this experiment, we evaluate FAIRS's ability to generalize to new datasets. For fair comparison with DEXTR, we conducted this experiment consistent with DEXTR's approach. That is, we report results with our model trained on both COCO and PASCAL, with both of their validation sets, as shown in table \ref{generalization_dataset}. FAIRS achieves a higher IoU in these evaluations and reaches an IoU of 85\% on COCO validation set when trained with COCO training set.

\begin{table}[b]
\parbox{.45\linewidth}{
\centering
\begin{tabular}{llcc}
\toprule
Train  & Test                                  & DEXTR  & Ours \\
\midrule
Pascal & COCO 2017 Val                & 80.1\% &  \textbf{81.76\%}    \\
COCO & COCO 2017 Val                   & 82.1\% &   \textbf{85\%}  \\
\midrule
COCO & Pascal Val                      & 87.8\% &     \textbf{90.6\%} \\
Pascal & Pascal Val                   & 91.5\% &    \textbf{92.2\%} \\
\bottomrule
\vspace{-1mm}
\end{tabular}
\caption{Evaluation of FAIRS's performance on unseen datasets.}
\vspace{0.85cm}
\label{generalization_dataset}
}
\hfill
\parbox{.45\linewidth}{
\centering
\begin{tabular}{lccc}
\toprule
Data     & FAIRS-WS & DEXTR & DELSE\\
\midrule
PASCAL  & 91.4\% & \textbf{91.5}\% & 91.3\% \\
COCO & \textbf{81.5}\% & 80.3\% & -- \\
\bottomrule
\vspace{-1mm}
\end{tabular}
\label{wsl_exp}
\caption{IoU results using FAIRS-COCO-Noisy on PASCAL and COCO datasets. We note that our method achieves IoUs on-par with DEXTR on both the datasets, despite never being trained on an actual ground truth.}
}
\end{table}

\subsection{Assisted Annotation -- Quality and Budget}
\label{assisted_annotation}
A key application of a tool such as FAIRS is instance segmentation from user inputs. In order to evaluate the quality of masks produced by FAIRS we conducted the following experiment. First, we used our PASCAL-trained model to produce annotations for COCO validation set, which is our hypothetical new dataset to be annotated. We refer to these FAIRS-produced annotations as COCO-GT-Noisy (GT=Ground Truth). Second, we train a new version of our model (not trained previously on any other segmentation dataset), using COCO validation set images, but instead of actual ground truth, we use generated labels COCO-GT-Noisy. We refer to this version of our model as FAIRS-WS (WS: weakly supervised). 

We, evaluate FAIRS-WS's performance on PASCAL, COCO, Berkeley, and GrabCut datasets. We note that FAIRS-WS was never trained on any segmentation ground truth data. Given the noise inherent in automatic labeling used to generate COCO-GT-Noisy, we highlight that extreme points input to the FAIRS-WS are noisy. Therefore, during training, FAIRS-WS has two difficulties to overcome: noisy annotations, and noisy user inputs (simulated extreme points from COCO-GT-Noisy). In this manner, we comprehensively test FAIRS's ability to create new annotations. We report the results with this model in table \ref{wsl_exp}, and additional results using FAIRS-WS have been shown in table \ref{sota_comparison}.

\noindent To elucidate these results, we highlight the following points:
\vspace{2mm}\\
\noindent\textbf{1.}\hspace{1mm} Using only the annotations generated from extreme points on \texttildelow36k objects, and with no segmentation pre-training with any ground truth data, our model achieves an IoU of 91.4\% on PASCAL validation set – on par with two state-of-the-art approaches [DELSE, DEXTR] that report similar result on PASCAL by training on \texttildelow25k objects with human-annotated ground truth annotations.
\vspace{2mm}\\
\noindent\textbf{2.}\hspace{1mm} We demonstrate that two versions of our model, one trained on PASCAL ground truth (\texttildelow10.5k images, \texttildelow25k objects), and the other trained on generated training data (5k images, \texttildelow36k objects), achieve nearly the same IoU. This demonstrates that annotations generated using FAIRS are effective for training fully-supervised segmentation models.

\noindent\textbf{3.}\hspace{1mm} Finally, we note that assuming 7.5 seconds as annotation time for extreme points \cite{Papadopoulos2017ExtremeAnnotation}, and \texttildelow2 minutes as a very conservative estimate of full annotation time (\cite{Papadopoulos2017ExtremeAnnotation} mention 55 seconds as median bounding box annotation time, and full annotation typically takes much longer), FAIRS-WS annotations (on COCO Validation set, \texttildelow35k objects) could be obtained \texttildelow11.4x faster than ground truth annotations (on PASCAL \texttildelow25k objects) at the expense of marginally lower performance compared to FAIRS.

\subsection{Interactive Object Segmentation}
\label{interactive_segmentation_result}
With FAIRS, user can start with 2 or more extreme points and add further positive or negative clicks. To demonstrate FAIRS's ability to encode corrective clicks for refinement, we trained our PASCAL model with additional positive and negative clicks. We simulated additional clicks by randomly sampling points within a distance of 15--60 pixels from the boundary of the mask to simulate refinement over false positive and false negative regions. We report results on PASCAL validation set for instances where IoU with 4 extreme points was less than 70\%, representing a scenario where the user might add a corrective click. The corrective click at test time was sampled by randomly sampling a point from the largest blob that contains either the false positives or false negatives. With the $5^{th}$ click added by random sampling for these hard samples, IoU improved by 6.1\% from 66.6\%, for relative gain of 9.2\%. We note that random sampling is not a fair representation of likely improvement, but represents a minimum gain achievable with the method. With an actual user click that is likely to be at a more conducive spot on the false positive or negative area, we expect the improvement to be greater. Lastly, in our experiments with only 3 extreme points, we observed that FAIRS was able to reach an IoU of 88.9\% on PASCAL, and 91.9\% on GrabCut. These results suggest that FAIRS can work well in an interactive segmentation mode, with a variety of click budgets.

\section{Conclusion}
In this study, we presented a novel scalable manner of incorporating cues from user-clicks, in a principled manner, in order to encode rich information for guiding a neural network towards the object of interest. Integrated with a dual attention module and a ResNet-101 backbone, we demonstrated through extensive experiments that FAIRS achieves its purpose of generating high quality data for fully supervised training, as evidenced by the results from FAIRS-WS. Finally, we demonstrated FAIRS's ability to handle <4 extreme points as well as corrective clicks in a single unified manner, enabled by soft focus generator. With these outcomes, we believe FAIRS can be an effective object segmentation tool.
\clearpage
%
%
\bibliographystyle{splncs04}
\bibliography{main}
\end{document}